\documentclass[letterpaper, 10 pt, conference]{ieeeconf}  

\IEEEoverridecommandlockouts                              

\usepackage{mathptmx} 
\usepackage{times} 
\usepackage{amsmath} 
\usepackage{amssymb}  
\usepackage{tabularx} 
\usepackage{graphicx}
\usepackage{comment}
\usepackage{mathtools}
\usepackage{upgreek}
\usepackage{gensymb}
\usepackage[noadjust]{cite}
\usepackage{placeins}
\usepackage{xcolor}
\usepackage{url}
\usepackage{mwe}
\usepackage{balance}
\usepackage{multirow}
\usepackage{amsmath}

\usepackage{dblfloatfix}
\newcolumntype{L}[1]{>{\raggedright\let\newline\\\arraybackslash\hspace{0pt}}m{#1}}
\newcolumntype{C}[1]{>{\centering\let\newline\\\arraybackslash\hspace{0pt}}m{#1}}
\newcolumntype{R}[1]{>{\raggedleft\let\newline\\\arraybackslash\hspace{0pt}}m{#1}}

\newcommand{\FK}{\text{\textit{FK}}}
\newcommand{\IK}{\text{\textit{IK}}}

\usepackage{mathtools}
\usepackage{wasysym}
\usepackage{bm}
\usepackage{bbm}

\usepackage{algpseudocode}
\usepackage[linesnumbered,ruled,vlined,noend]{algorithm2e}
\SetKwRepeat{Do}{do}{while}

\DeclarePairedDelimiter\abs{\lvert}{\rvert}%
\DeclarePairedDelimiter\norm{\lVert}{\rVert}%

\makeatletter
\let\oldabs\abs
\def\abs{\@ifstar{\oldabs}{\oldabs*}}
\let\oldnorm\norm
\def\norm{\@ifstar{\oldnorm}{\oldnorm*}}
\makeatother

\title{\LARGE \bf
Physical Twins: Accelerating and Enabling Robot Learning with Phantom Platforms
}

\author{Elizabeth Peiros$^{1\dag}$, Derek Chen$^{1\dag}$, Lucas Yager$^{2\dag}$,  Edward Lee$^2$, Kevin Wang$^2$, Zoe Samuels$^2$, \\Michael C. Yip$^1$ \IEEEmembership{Senior Member, IEEE}
\thanks{$^\dag$ denotes equal contribution}%
\thanks{$^1$ Derek, Elizabeth, and M.C. Yip are with the Electrical and Computer Engineering Department, University of California, San Diego, La Jolla, CA 92093 USA. {\tt\footnotesize\{dec004, epeiros, yip\}@ucsd.edu}}%
\thanks{$^2$ Lucas Yager, Edward Lee, Kevin Wang, and Zoe Samuels are with the Mechanical and Aerospace Engineering Department, University of California, San Diego, La Jolla, CA 92093 USA. {\tt\footnotesize \{lyager, edl003, kew054, zsamuels\}@ucsd.edu}}%
}

\begin{document}

\maketitle
\thispagestyle{empty}
\pagestyle{empty}


\begin{abstract}
Improvements in human-robot physical interaction (pHRI) can have major implications for physical therapy, search and rescue, and telemedicine. However, a major challenge concerns human constraints and safety in human-robot physical experiments. Concerns about human studies also include repeatability, scalability, and participant diversity. To conduct such experiments, an IRB and willing human participants are required. In this work, we present an improved phantom device, a physical twin, that enables real-world RL-type testing for physically interactive algorithms. The new device not only replicates the ball-and-socket motion of the shoulder but also renders scapular and protraction/retraction motions. The experiments showcase the device's ability to render multiple 3D joint limits and demonstrate a Franka Panda arm sensing limits and interacting with the device as an arm for ADLs (activities of daily living) and pHRI tasks.
\end{abstract}

\section{Introduction}
\vspace{-1mm} 

The primary cause of long-term disabilities in the United States is strokes, which affect more than 700,000 Americans annually. Among these stroke patients, 69\% experience motor impairment in their upper extremities \cite{urton2007systematic}. This is just one of many medical events or disabilities that could require home care and assistance with daily living. Many activities of daily living (ADLs) require the use of the upper extremity, such as washing hair, getting dressed, and weight shifting. Those who cannot perform these actions themselves rely on others, such as a loved one or a nurse, to provide care. Some of the challenges for these caregivers include the physical demands of the job, lifting and moving a heavy body with little to no assistance from the care recipient. Taking on this responsibility can lead to burnout \cite{gerain2021informal,froese2009emotional}. A nurse can take on some of the care tasks, yet nurses, too, are susceptible to limitations in strength for moving patients and burnout due to high demand and nursing shortages\cite{shah2021prevalence,reith2018burnout,haddad2023nursing}. 

Physical robotic assistance for moving patients would reduce this caregiver burden. There have been significant advances in enabling robotic platforms to perform physically demanding tasks, including helping people dress \cite{wang2023one} and assisting people who cannot shift weight in bed by grasping and pulling bed sheets \cite{1241924}. Specific medical and outpatient care work has even demonstrated robots' ability to perform teleoperated and automated heart listening \cite{10089533}, automated diagnosis of arthritis using ultrasound \cite{frederiksen2022ultrasound}, and a variety of other tasks \cite{atar2025humanoids}. In search and rescue, steps have been taken to grasp and manipulate limbs with biomechanics considerations \cite{zhao2024high,peiros2023finding}. However, research on the direct movement and maneuvering of patients remains limited. Safety concerns about conducting human trials are a driving limitation.

\begin{figure}
    \vspace{1mm}
    \centering
    \includegraphics[trim=15cm 0cm 58cm 0cm, clip, width=1\linewidth]{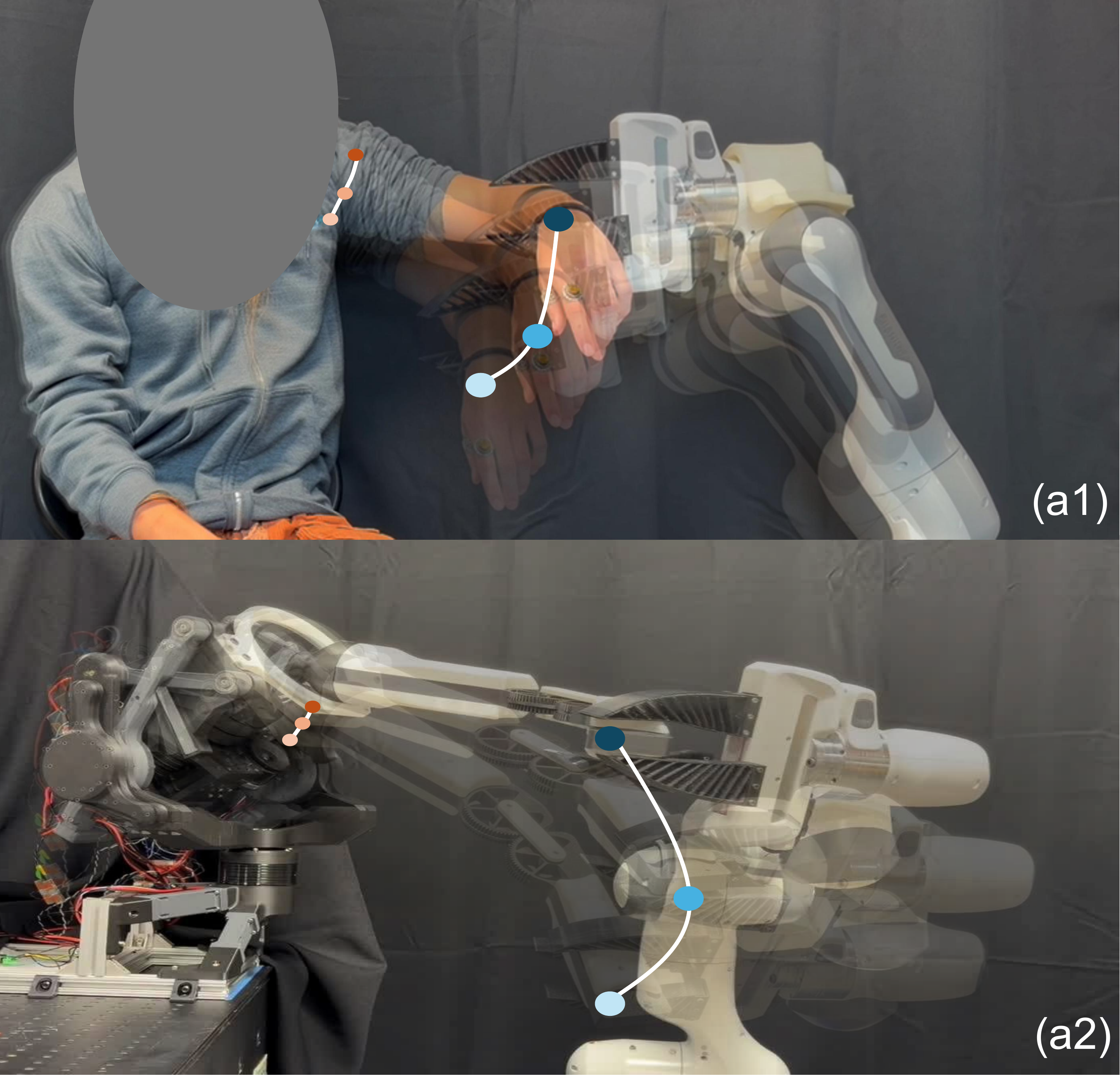}
    \caption{The embodiment of the physical twin showing the full mechanical design and two sub-components. The image on the right shows both a human and the physical twin being moved by a robotic manipulator.}
    \label{fig:cover_fig}
    \vspace{-4mm}
\end{figure}

For research in physical therapy robotics, physical robotic interaction with patients is required. In most cases, the method of choice is to use an impedance controller\cite{tejwani2024demonstrating}. This can work against the robot, reducing its strength in tasks such as transfers and limb maneuvering. In the context of safety, deep biomechanical modeling combined with discomfort maps has been used to estimate tendon strain in able-bodied shoulder rehabilitation patients \cite{balvert2022enabling, fang2023human, beck2023real, ravenberg2024creating}. These new steps are essential for unlocking real-world patient care and maneuvering. However, these methods are less adaptable and require significant effort to set up to align the model with the patient. 


Most models are generic or averaged from data collected from many individuals, providing an "average user" but not an accurate one. Therefore, it is mandatory to manually fit the model to the individual's parameters. When considering the high variability of disability, these high-fidelity models still fall short. It is imperative that robotics training data include individuals with conditions such as muscular dystrophy (MD) or spinal muscular atrophy (SMA), in which posture and movement patterns are highly differentiated from those of able-bodied people, joint ranges of motion fall well below average, and joint movements are inconsistent. There is a clear gap between current robotics capabilities and the technology that could best serve the disabled and older adult community.


We propose a different approach to training robotic assistance, in which robots can learn intuitions similar to those of human caregivers, rather than using complex, high-fidelity models unsuitable for real-time tasks in homes or overly compliant controllers that cannot contextualize resistive forces. As caregivers, humans lack detailed insight into the musculoskeletal makeup of the shoulder joint, yet are capable of performing safe tasks. Human force perception through touch, specifically resistance, provides all the necessary feedback. This informs them of the end ranges of motion without requiring an internal model, thereby enabling rapid, dynamic, real-time changes in action. Any useful physical assistance requires real-world online learning and greater consideration of user variation, in addition to close human models. 

In this work, we introduce a physical twin, akin to a phantom used to train medical personnel, inspired by the original concept in \cite{peiros2025haptic}. The physical twin is a physical replication of the human patient, providing a haptic environment that simulates joint limits, such as those of the human shoulder. The advantages of such a platform include: removing the safety risk of initial testing on human subjects, an infinite set of users capable of rendering an infinite set of virtual joint limits, and internal ground-truth measurements unattainable in human subjects. We achieve this through the design of a bio-realistic shoulder joint, which includes the 5 degrees of freedom of the scapula and ball-and-socket motion. Motors are integrated and programmed to provide virtual tendon and soft-tissue forces at a safe scale, ensuring safe testing and access to inclusive user data.

\section{Related Works}

There are three bodies of work that correspond to the task of developing safe robotics systems for physical human-robot interaction (pHRI): (1) applications in moving humans and monitoring vital health signs, (2) safety considerations regarding physiological limits, and (3) mechanical designs of humanoid and anthropomorphic arms. 

\subsubsection{Physical human-robot interaction}
Robotic systems have increasingly been deployed for pHRI tasks related to moving people and health monitoring. Researchers have explored the use of robots for transfer tasks \cite{atar2025humanoids}, dressing clothes \cite{wang2023one}, and weight shifting \cite{1241924}. Not only that, but they have been used for automating heart listening \cite{10089533} and monitoring vital health signs \cite{koceska2019telemedicine}. Further applications extend to human-robot collaboration for physical therapy \cite{tejwani2024demonstrating} and search and rescue operations \cite{peiros2023finding}. However, the physical interaction in these tasks introduces risks. For example, studies show that bone fractures and skin and soft tissue damage from lower limb exoskeletons are a known risk of device use \cite {he2017risk}.

\subsubsection{Safety}
Ensuring safety is crucial for robotic systems to prevent harm to people who are interacting with them. Critical safety considerations include respecting the end ranges of motion in joints to prevent injury \cite{zhao2024high} and reducing the risk of dislocations within the joint spaces of the human body. To prevent injury and respect physiological limits, collaborative robots (cobots) have utilized impedance control \cite{madan2024rabbit} or constrained optimization methods \cite{sun2024force} for compliant interaction. Further methods also employ Model Predictive Control (MPC) \cite{sacks2024deep} to predict states and optimize torque, ensuring safe, seamless operation.

\subsubsection{Mechanical Designs}
Current mechanical designs for humanoid and anthropomorphic arms strive to replicate human capabilities but face limitations \cite{tong2024advancements}. To make the robots easier to build, designers often use simple serially-linked mechanisms rather than perfectly copying the complex structure of human anatomy \cite{ceccarelli2020parallel}. The biggest problem is usually the shoulder. While a human shoulder is a flexible ball-and-socket joint that can roll in many directions, most robots rely on fixed rotations \cite{peiros2025haptic}. This rigid design limits the robot's movement, preventing it from truly replicating the natural motion of human shoulders \cite{cao2025enhancing}. While there have been attempts to solve this with more anthropomorphic designs, they usually end up being very bulky and expensive to manufacture \cite{jiang2026bionic}.

\section{Methods}

Our physical twin design features a mechanical mechanism that emulates the complex anatomy of the shoulder, with attention to realistic grasping and maneuvering points: the wrist, arm, and armpit, typically used for lifting, dressing, and weight shifting. The device provides the robot manipulator with forces that are sensible at a scale safe for the lab environment and supports an infinite number of individual user profiles, enabling proper testing of the true safety of a robotically assisted task. To that end, we created a mechanical design with all 5 degrees of freedom in the shoulder joint with a maximum range similar to that of an able-bodied individual, an electrical system to sense and provide resistive torques, and software that can take a unique range of motion and deploy that to the kinematics of our system with the use of haptic control.

\subsection{biomechanics}
In this paper, the shoulder joint will be described as a combination of multiple joints to illustrate its anatomical motion and its replication in the mechanical design.

\subsubsection{Relevant Anatomy}
\label{sec: bio_anatomy}

Anatomically the shoulder is comprised of 4 joints: (1) the glenohumeral (GH) joint(ball and socket articulation between the head of the humerus and the glenoid cavity of the scapula), (2) the acromioclavicular (AC) joint (where the clavicle meets the acromion of the scapula), (3) the ternoclavicular (SC) joint (where the clavicle meets the chest bone [sternum]), and (4) the scapulothoracic joint (where the scapula meets with the ribs at the back of the chest). The anatomy can be visually seen in Fig \ref{fig: joint2mech} (A). 

To simplify the complex anatomy, these joints will be categorized into shoulder and scapular motion. The \textit{shoulder motion} will capture the motion of the glenohumeral joint, and \textit{scapular motion} will include the AC, SC, and scapulothoracic, which control the translation and rotation of the glenoid cavity.

\subsubsection{Capturing Human Motion}
\label{sec: bio_hum_motion}

\begin{figure}[t!]
\vspace{2.5mm}
    \centering
    \includegraphics[trim={0cm 23cm 0cm 0cm},clip,width=1\linewidth]{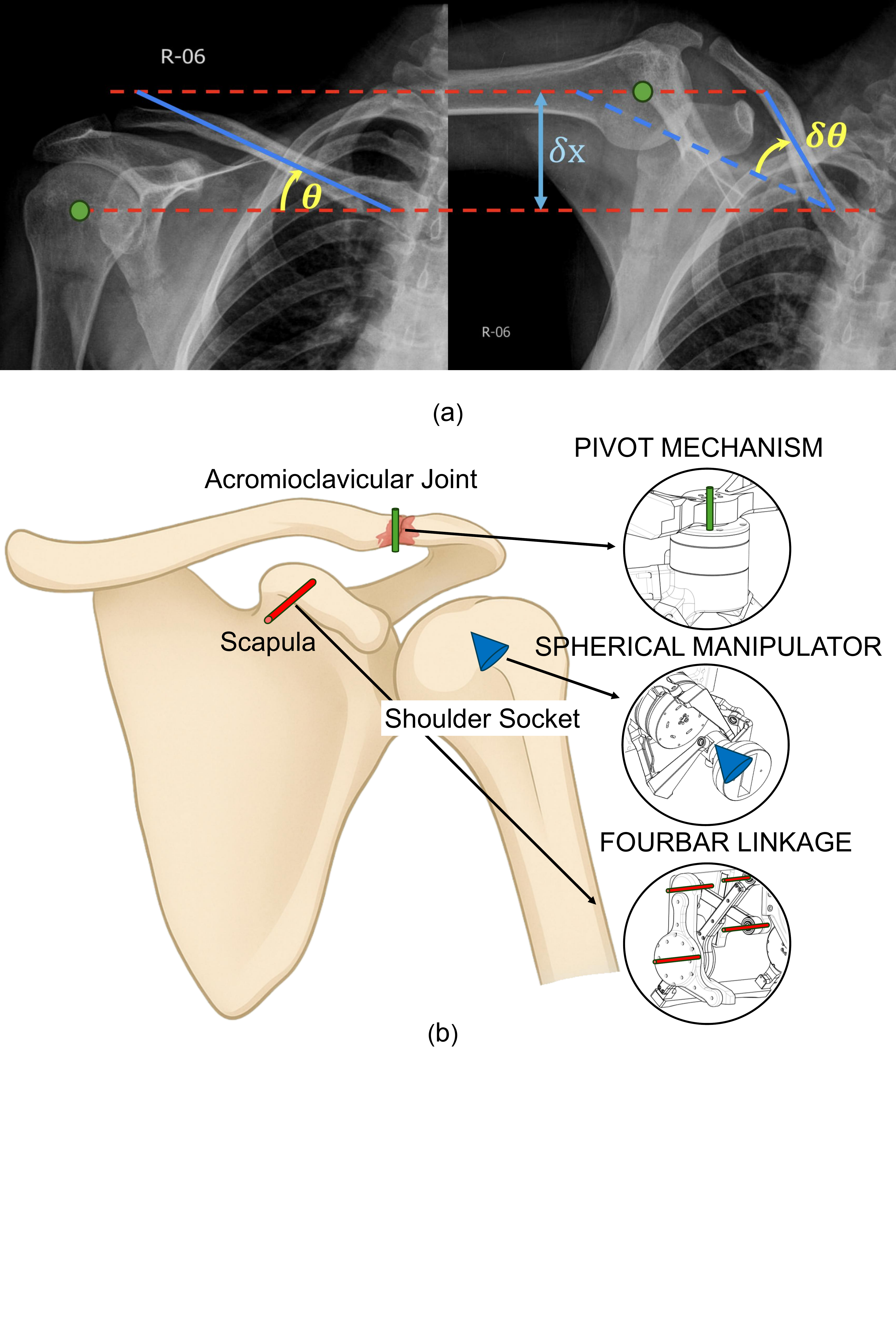}
    \caption{Top image, Fig (a), highlights the rotation of the clavicle bone when x-rayed. The labeling shows both the angular change and the translational motion of the shoulder socket. Bottom image, Fig (b), shows the translation of joint motion into a mechanical mechanism. From top to bottom the acromioclavicular joint is transformed into a pivot joint, the shoulder socket is transformed into a spherical manipulator, and the scapular motion is re-imagined as a 4-bar mechanism.}
    \label{fig: joint2mech}
\end{figure}

To increase biomechanical accuracy, this design includes both scapular and glenohumeral motion to capture a more realistic shoulder motion. While ball-and-socket joints are a more straightforward design to replicate, the scapular joint is far more complex. It is more similar to a 3D linkage system, in which the clavicle, scapula, and rib cage function as bars. A visual of the motion parameters of interest can be seen in Fig \ref{fig: joint2mech} (a), where the shoulder is x-rayed and annotated to show the rotation and translation of the shoulder socket. 

\subsubsection{Range of Motion}
\label{sec: bio_range}

Past work capturing shoulder movement reports an active range of glenohumeral elevation of $90 \degree$, while the scapular elevation is capped at $60 \degree$ \cite{cailliet1981shoulder}. Translationally, this leads to an upward and downward rise of the shoulder socket pivot. The maximum vertical scapular translation observed was found to be 6.24 cm \cite{Baumgarten2012-xk}. Similarly, for external rotation, the scapular protraction/retraction reaches $40 \degree$, while the internal and external rotation of the glenohumeral is reported to be $160\degree$ total \cite{luttgens1982kinesiology}.

\begin{figure*}[t!]
    \vspace{2.5mm}
    \centering
    \includegraphics[width=1\linewidth]{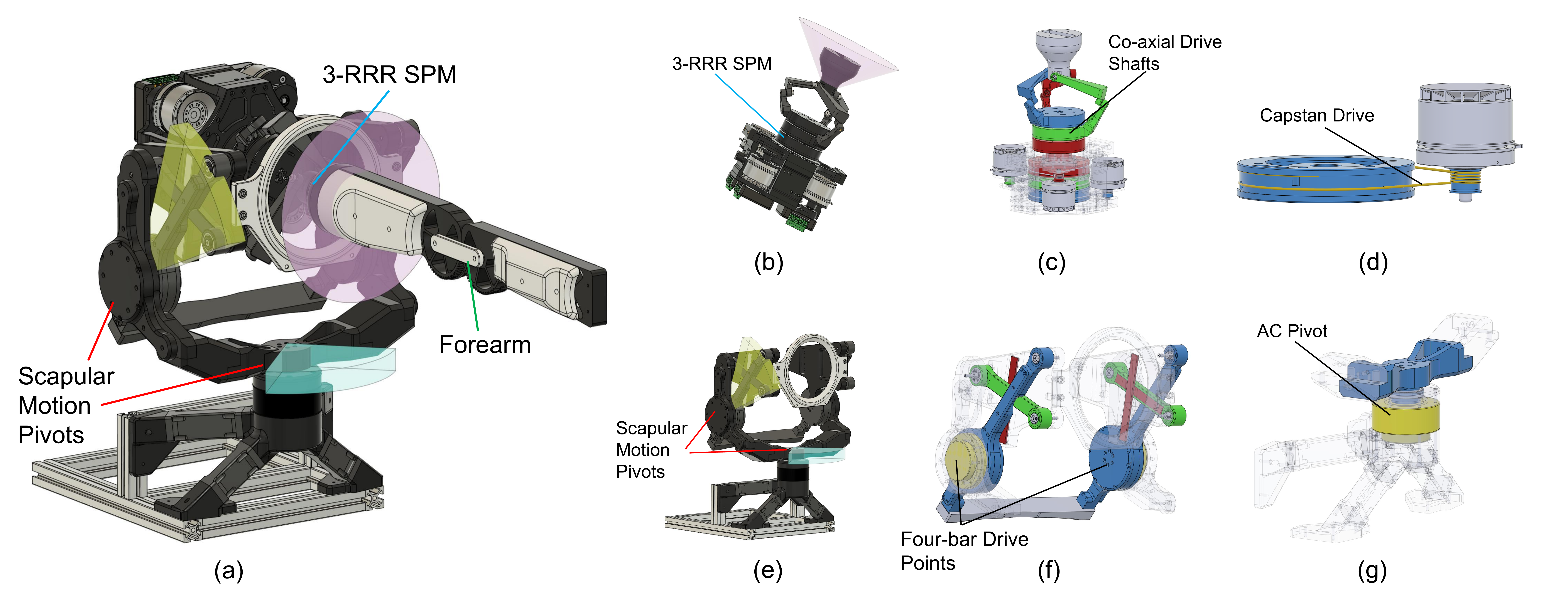}
    \caption{(a) Model of the haptic shoulder with each components range of motion demonstrated in yellow, pink, and cyan (b) The Spherical Parallel Manipulator (SPM) Mechanism separated out (c) The concentric shaft coupling of the SPM with linked linkages demonstrated with the same color (d) The capstan drive mechanism for the SPM with steel string highlighted in yellow (e) The four-bar and pivot mechanisms separated out (f) Individual components of the four-bar mechanics with yellow being the driving motor, green and blue being the linkages, and red being the PTFE slides (g) Components of the pivot with the motor being highlighted in yellow}
    \label{fig:mech_design_temp}
    \vspace{-4mm}
\end{figure*}

\begin{table}[t!]
\vspace{2.5mm}
\centering
\caption{Comparison of Maximum Shoulder Joint Range of Motion}
\label{tab:joint_limits}
\begin{tabular}{lll}
\hline
\textbf{Joint Movement} & \textbf{Literature Limit} & \textbf{Our Design} \\ \hline
GH elevation & $90^\circ$ & $94^\circ$ \\
GH internal/external rotation & $160^\circ$ & $400^\circ$ \\ 
Scapular protraction/retraction & $40^\circ$ & $140^\circ$ \\
Scapular elevation & $60^\circ$ & $114^\circ$ \\
Vertical scapular translation & 6.24 cm & 10.75 cm \\ \hline
\end{tabular}
\vspace{-1mm}
\end{table}

\subsection{Mechanical design}
To accurately capture the maximum range of human shoulder motion determined from literature and render forces noticeable to a robotic manipulator (in this case a Franka Panda), the shoulder incorporates a design that mimics the human physical shoulder using a 4-bar mechanism and pivot to capture scapular motion and an spherical parallel manipulator to capture glenohumeral motion as seen in Fig \ref{fig: joint2mech}(b). A summary comparing the joint limits found in literature to joint limits of the physical twin can be found in Table \ref{tab:joint_limits}

\subsubsection{Force Rendering Detectable Within Franka Panda Limitations}
Baseline motor calculations were performed using a worst-case scenario that a single motor provides all haptic feedback for the spherical parallel manipulator, with the arm fully extended. 
The minimum force that can be perceived by the panda must be rendered at the grasp point; therefore, the minimum input torque is governed by \ref{eq:torque_min}:

\vspace{-3mm}
\begin{equation}
    \tau_{a, min} = (NF \cdot N_{joints} \cdot SNR_{factor}) \cdot (H + U)
    \label{eq:torque_min}
\end{equation}

Where $NF$ is the intrinsic noise floor, $N_{joints}$ is the maximum number of joints the input force can be distributed across, $SNR_{factor}$ is the desired signal-to-noise ratio, and $(H + U)$ represents the total effective lever arm length (humeral and ulna segments). The effective reduction of the spherical parallel manipulator is taken to be 1:1. The specific numbers utilized for the calculation are summarized in Table \ref{tab:haptic_motor_specs}.

\begin{table}[t!]
\centering
\caption{System parameters for the Franka Panda force detection}
\label{tab:haptic_motor_specs}
\begin{tabular}{lcll}
\hline
\textbf{Parameter} & \textbf{Symbol} & \textbf{Value} & \textbf{Units} \\ \hline
Noise Floor & $NF$ & 0.035 & N \\
Joint Count & $N_{joints}$ & 7 & -- \\
SNR Factor & $SNR_{factor}$ & [5] & -- \\
Total Length & $H+U$ & [0.57] & m \\ 
Min Input Torque & $tau_{a, min}$ & [0.698] & Nm \\\hline
\end{tabular}
\vspace{-1mm}
\end{table}

\subsubsection{Scapular Motion Mechanism }
The 4 Bar mechanism employs an inverted design of the linkage elements, seen in Figure \ref{fig:mech_design_temp} (a, e, f), that captures scapular motion. This design models the vertical scapular translation as the scapula moves up and down, designed to be 10.75 cm in total, and the rotational angle change of scapular motion, designed to be 114\degree. Combined with the pivot joint, seen in Figure \ref{fig:mech_design_temp} (g), which provides a rotation about the vertical axis of 140\degree and approximates AC rotation, these mechanisms cover the range of motion of the human scapula as seen in Table \ref{tab:joint_limits}.

\subsubsection{Spherical Parallel Manipulator (SPM) }
Glenohumeral motion is emulated using a coaxial spherical parallel manipulator. The specific mechanism employs three concentric rotational axes (3-RRR), driven by a 6:1 reduced-capstan drive shown in Figure \ref{fig:mech_design_temp} (a, b, c, d). The concentric shafts drive three corresponding arms that intersect at the center of spherical rotation with an angle of 15 degrees. Combined, the mechanism can cover a combined 94\degree, and 400\degree of rotation, covering more than the typical able-bodied range of glenohumeral rotation found in Table \ref{tab:joint_limits}.

\begin{figure*}
\vspace{2.5mm}
    \centering
    \includegraphics[trim={0cm 0mm 0cm 11cm},clip,width=.99\linewidth]{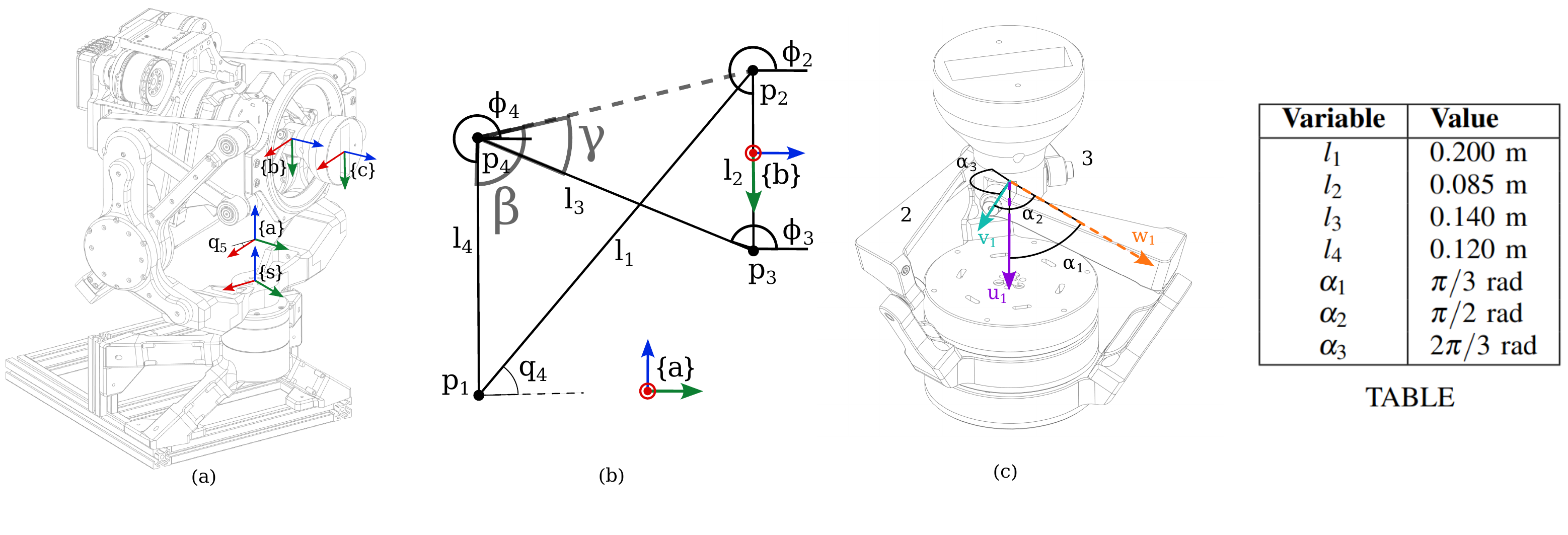}
    \vspace{-6mm}
    \caption{(a) Frames of the haptic shoulder, frame $\mathcal{C}$ is shown offset along the $z_b$ axis of $\mathcal{B}$ for visual clarity as the origins of $\mathcal{B}$ and $\mathcal{C}$ are coincident. (b) Diagram of the 4-bar mechanism with lengths, $l_{1:4}$, and joints at points $p_{1:4}$ and the $\mathcal{A}$ and $\mathcal{B}$ frames. (c) Agile eye with $\hat{\mathbf{u}}$, $\hat{\mathbf{v}}$, and $\hat{\mathbf{w}}$ vectors shown with arms 2 and 3 labeled. The table shows the mechanical constants (lengths) of SPM and 4-bar}
    \label{fig:Kinematic_Frames}
    \vspace{-4mm}
\end{figure*}

\subsection{Kinematics}
The kinematics of the device will be defined using 3 sequential reference frames [$\mathcal{A}$, $\mathcal{B}$, $\mathcal{C}$] relative to the fixed frame $\mathcal{S}$. In sequence, the $\mathcal{S}$ frame's origin is located in the joint center of the protraction/retraction motor. The $\mathcal{A}$ frame's origin is translated vertically to align with the 4-bar base, and rotates with the device about the $\hat{s_z}$ axis with positive sense. The $\mathcal{B}$ frame's origin is coincident with the SPM joint center and has a rotation about $\hat{a_x}$ with positive sense, the normal to the plane of the 4-bar. Finally, frame $\mathcal{C}$'s origin is at the end effector of the SPM. All the frames can be seen in Fig \ref{fig:Kinematic_Frames}(a)

\subsubsection{Kinematics of Spherical Parallel Manipulator (SPM)}

As this is a well-studied mechanism, former designers have derived the kinematic chain for the SPM \cite{tursynbek2019computation}.
Proper design requires all rotation axes on the joints to intersect, creating a single joint center. The forward kinematics can be solved by exploiting two geometric constraints: a shared center of rotation (COR) and the known angle between the joint axes of each rigid link. For each arm, the vectors $\hat{\mathbf{u}}$, $\hat{\mathbf{v}}$, and $\hat{\mathbf{w}}$ will be defined along the axis of rotation for each joint. The legs are numbered clockwise about $\hat{c_z}$ in the neutral position. The $\hat{u_i}$ vectors point from the shared joint COR to the revolute joint of the proximal link. The $\hat{w_i}$ vectors point to the joints that connect the proximal and distal joints, and the $\hat{v_i}$ vectors point to the joints that connect the distal links to the platform. $\alpha_1$ is the fixed angle between the $u$ and $w$ vectors and $\alpha_2$ is the fixed angle between the $v$ and $w$ vectors. Because the SPM is coaxial $\hat{\mathbf{u_i}}=[0\;\;\;0\;-1]^T$ in the $\mathcal{C}$ frame and $\hat{\mathbf{w_i}} = [\sin(\eta_i - q_i)\sin\alpha_1\;\; \cos(\eta_i - q_i)\sin\alpha_1\;\; -\cos\alpha_1]^T$ where,
$\eta_i = 2(i-1)\pi/3$ $, i =1,2,3$

We can then derive 9 equations that can be solved to obtain the vectors $v_{1:3}$. The first 3 being the magnitude constraint that all are unit vectors $\|\mathbf{v}_i\| = 1$ and the other 6:

\vspace{-3mm}
\[
\begin{cases}
\hat{\mathbf{w}}_i \cdot \hat{\mathbf{v}}_i = \cos\alpha_2, & i = 1,2,3, \\[3pt]
\hat{\mathbf{v}}_i \cdot \hat{\mathbf{v}}_j = \cos\alpha_3, & i,j = 1,2,3,\; i \ne j
\end{cases}
\]

Vectors are then formed to create the homogeneous transform $T_{bc}$: $\hat{x_c}=v_1$, $\hat{z_c}=v_2 \times v_1$, and $\hat{y_c}=\hat{z_c} \times \hat{x_c}$.

Based on previous studies of the SPM that derive the Jacobian mapping end effector rotational velocities to joint velocity \cite{cruz2021kinematic}, we can find $J_{SPM}$ that maps from joint velocity to end effector rotational velocity.
\begin{equation}
J_{SPM} = \begin{bmatrix}
\mathbf{j}_1 & \mathbf{j}_2 & \mathbf{j}_3
\end{bmatrix}^{-T},
\qquad
\mathbf{j}_i =
\frac{(\mathbf{w}_i \times \mathbf{v}_i)^{T}}
{(\mathbf{-u}_i \times \mathbf{w}_i)\cdot \mathbf{v}_i}.
\end{equation}

Inverse kinematics can be calculated using similar convenient vectors to get the proximal link angles, $q_{1:3}$, for a specified end effector configuration $T_{bc}$. The $\hat{\mathbf{v}}_i$ vectors are derived from the construction of $T_{bc}$. Then, using the quadratic equation below, $q_i$ can be derived.
\vspace{-3mm}
\begin{align*}
A_i& = -v_{iy} \sin \alpha_1 - v_{iz} \cos \alpha_1 - \cos \alpha_2,\;\;\; B_i = v_{ix} \sin \alpha_1,\\
C_i& = v_{iy} \sin \alpha_1 - v_{iz} \cos \alpha_1 - \cos \alpha_2.
\end{align*}

When $B_i^2-A_iC_i<0$, there is no inverse kinematic solution for the given $T_{bc}$ \cite{bai2010optimum}.

\begin{equation}
A_iT_i^2+2B_iT_i+C_i=0, \;\;\; T_i=\tan(\frac{\eta_i-q_i}{2})
\end{equation}

Next is to define $T_{ab}$, the transform between frame $\mathcal{A}$ and frame $\mathcal{B}$. This transform will have both translational and rotational components, even though it is a single degree of freedom. Points $p_i$, with position vector $\mathbf{p}_i$, are labeled on each of the revolute joints along with lengths $l_i$ as seen in Fig \ref{fig:Kinematic_Frames}(b). The points $p_1$ and $p_4$ are fixed in frame $\mathcal{A}$. The planar geometry constrains the motion to use the law of cosines to calculate $\mathbf{p}_2$ and $\mathbf{p}_3$, the positions of $p_2$ and $p_3$ relative to frame $\mathcal{A}$'s origin.

\vspace{-3mm}
\begin{align}
 \gamma &= \arccos(\frac{||\mathbf{p}_2-\mathbf{p}_4||^2+l_3^2+l_2^2}{2l_3||\mathbf{p}_2-\mathbf{p}_4||})\\
    \beta &= \arctan2(p_{2z}-p_{4z},p_{2y}-p_{4y})\\
    \mathbf{p}_3&=\mathbf{p}_4+(0,l_3\cos(\beta-\gamma),l_3\sin(\beta-\gamma)) 
\end{align}

Then the origin of $\mathcal{B}$ can be found. $\mathbf{p}_{ab}=\frac{\mathbf{p}_2+\mathbf{p}_3}{2}$. The direction vectors of frame $\mathcal{B}$ expressed in frame $\mathcal{A}$ can now be found as follows: $\mathbf{\hat{x}}_b=\mathbf{\hat{x}}_a$
,
$\mathbf{\hat{y}}_b=\frac{\mathbf{p}_3-\mathbf{p}_2}{||\mathbf{p}_3-\mathbf{p}_2||}$
,
$\mathbf{\hat{z}}_b=\mathbf{\hat{x}}_b \times \mathbf{\hat{y}}_b$. 
The transformation matrix $T_{ab}$ is then a homogeneous transform using these components. The Jacobian of the 4-bar is:

\begin{equation}
    \mathbf{J_{scapula}}=\begin{bmatrix}
    \mathbf{J_\omega} \\ \dot{\mathbf{p_{ab}}} + [\mathbf{p_{ab}}]\mathbf{J_\omega}  
\end{bmatrix}
\end{equation}

\vspace{-2mm}
\begin{equation}
    \mathbf{J_{\omega,Scapula}}=\begin{bmatrix}
    \frac{l_1\sin(q_4-\phi_3)}{l_2\sin(\phi_3-\phi_2)} & 0 &0
\end{bmatrix}^T
\end{equation}

\vspace{-4mm}
\begin{multline}   
\mathbf{\dot{p}_{ab}}=l_1\begin{bmatrix}
    0 & -\sin(q_4) & \cos(q_4) 
\end{bmatrix}^T\\
+\frac{l_2 J_{\omega_x}}{2}\begin{bmatrix}
    0 & -\sin(\phi_2) & \cos(\phi_2)
\end{bmatrix}^T
\end{multline}

Finally, $T_{sa}$ retraction and protraction of the shoulder is defined by a simple rotation about the $\mathbf{\hat{z}}_s$ = $\mathbf{\hat{z}}_a$ axis by $q_5$ with positive sense. The origin of $\mathcal{A}$ is placed .0822m above the $\mathcal{S}$ frame in the $\mathbf{\hat{z}}_s$ direction to align with the vertical scapular center of rotation. The Jacobian of protraction and retraction is then $\mathbf{J_{PR}}= [0\;\;0\;\;1\;\;0\;\;0\;\;0]$

\subsubsection{Static Force Production}
\label{sec: forceProd}
The torques required to produce a desired wrench in the $\mathcal{S}$ frame, $\mathbf{\mathcal{F}_s}$, can be calculated using $\mathbf{\tau_m}=J_s(q)^T \mathbf{\mathcal{F}_s}$ where $J_s$ is the full Jacobian of the shoulder in the $\mathcal{S}$ frame:
\vspace{-2mm}
\begin{equation}
    J_s=\begin{bmatrix}
    [Ad_{T_{sb}}]J_{SPM} & [Ad_{T_{sa}}]J_{Scapula} & J_{PR} 
\end{bmatrix}
\end{equation}

\subsection{System Design and Communication} 

The device incorporates three Maxon BLDC motors (PN: 647697) with integrated hall sensors and 3 Cubemars AKA60-6 motors. Each maxon motor has an ODrive Pro controller operating in Field Oriented Control. All motors are connected to a 48 V DC rail with a current limit set to 40 A on the power supply.

System communication is handled through an RS485 CAN hat at a bitrate of 1,000,000 bits per second. The CAN lines are daisy-chained across all ODrive Pro controllers and Cubemars motors on a shared bus. Each Odrive controller operates in Ramped Torque Control Mode with a firmware-enforced torque limit of 0.6 Nm to prevent mechanical overload. The Cubemars motors are run using the MIT control mode.

\subsection{Software}

The limits of the shoulder joint are defined using a sinus/reach cone that best captures its complexity. The central point of the cone coincides with the center of rotation of the scapular motion, defining the combined range of all three mechanical components into a single joint range of motion shown in \cite{wilhelms2001fast, peiros2025haptic}.

\subsubsection{Defining Joint Limits}

In this work, the cone is defined via the clinical extrinsic angles, and the surface normals $\boldsymbol{S}_{i} \in \mathbb{R}^3$ and $\boldsymbol{B}_{i} \in \mathbb{R}^3$ are defined via interpolated points $\boldsymbol{P}_{i} \in \mathbb{R}^3$ defining the edges of the reach cone and the visible point $\boldsymbol{V} \in \mathbb{R}^3$:
\vspace{-3mm}
\begin{align}
    \boldsymbol{B}_i = \boldsymbol{P}_i \times \boldsymbol{P}_{i+1}, \;\;\;\;
    \boldsymbol{S}_i = \boldsymbol{V} \times \boldsymbol{P}_{i}
\end{align}

\subsubsection{Calculating Error}

In this design, the clinical frame corresponds to the frame $\mathcal{S}$, however, transforming clinically measured angles into Euler angle sequences or quaternian mapping typically requires a rotational transformation similar to
\cite{wu2005isb, wren2007simple, chenier2022interpreting}. 

A more comprehensive calculation of error can be found in \cite{peiros2025haptic}. The equations below show how to compute the axis-angle result where $\theta_e$ is the angle error, $\boldsymbol{\lambda} \in \mathbb{R}^3$ defines the axis of rotation, $\boldsymbol{L}$ a vector aligned with the humeral bone, and $\boldsymbol{L}_B$ is the humeral bone projected onto the outer surface of the joint limits:

\vspace{-4mm}
\begin{align}
    \theta_e = cos^{-1}(\boldsymbol{L} \cdot \boldsymbol{L}_B), \;\;\;
    \boldsymbol{\lambda} = \boldsymbol{L} \times \boldsymbol{L}_B
\end{align}

\subsubsection{Haptics}

Aligning the system kinematics with the human shoulder kinematics allows the device to sense human shoulder pose and provide appropriate resistance forces relative to the loaded phantom human limits derived from user data.

\begin{equation}
\begin{cases} 
    \boldsymbol{\tau}_{s} = - (k_p \theta_e)\boldsymbol{\lambda} \;\;\;\theta \notin \boldsymbol{\Theta}_{free}\\
    \boldsymbol{\tau}_{s} = 0 \;\;\;\;\;\;\;\;\;\;\;\;\;\theta \in \boldsymbol{\Theta}_{free}
\end{cases}
\end{equation}

Where $\boldsymbol{\tau}_{s} \in \mathbb{R}^3$ is the desired wrench (torques only) at the end effector, $k_p \in \mathbb{R}$ is the torsional spring constant.

\subsubsection{Calculate Motor Torques}
Using the Jacobian described in sec \ref{sec: forceProd}, the motor torques can be calculated to produce the appropriate resistive forces 

\begin{equation}
    \boldsymbol{\tau}_{m} = ([Ad_{T_{sb}}]J_{SPM}(\boldsymbol{q}))^{T}\begin{bmatrix}
    \boldsymbol{\tau}_s \\ \boldsymbol{0}
\end{bmatrix}
\end{equation}


\subsubsection{Feedforward Gravity Compensation}

Gravity compensation was applied assuming a 1.86 kg mass at .2 m in the $-\hat z_b$ direction. The gravity compensation uses between 2.2 Nm and 6.6 Nm of the total 8 Nm of available torque throughout the scapular range of motion.

\begin{figure}[t!]
\vspace{2.5mm}
    \centering
    \includegraphics[width=0.99\linewidth]{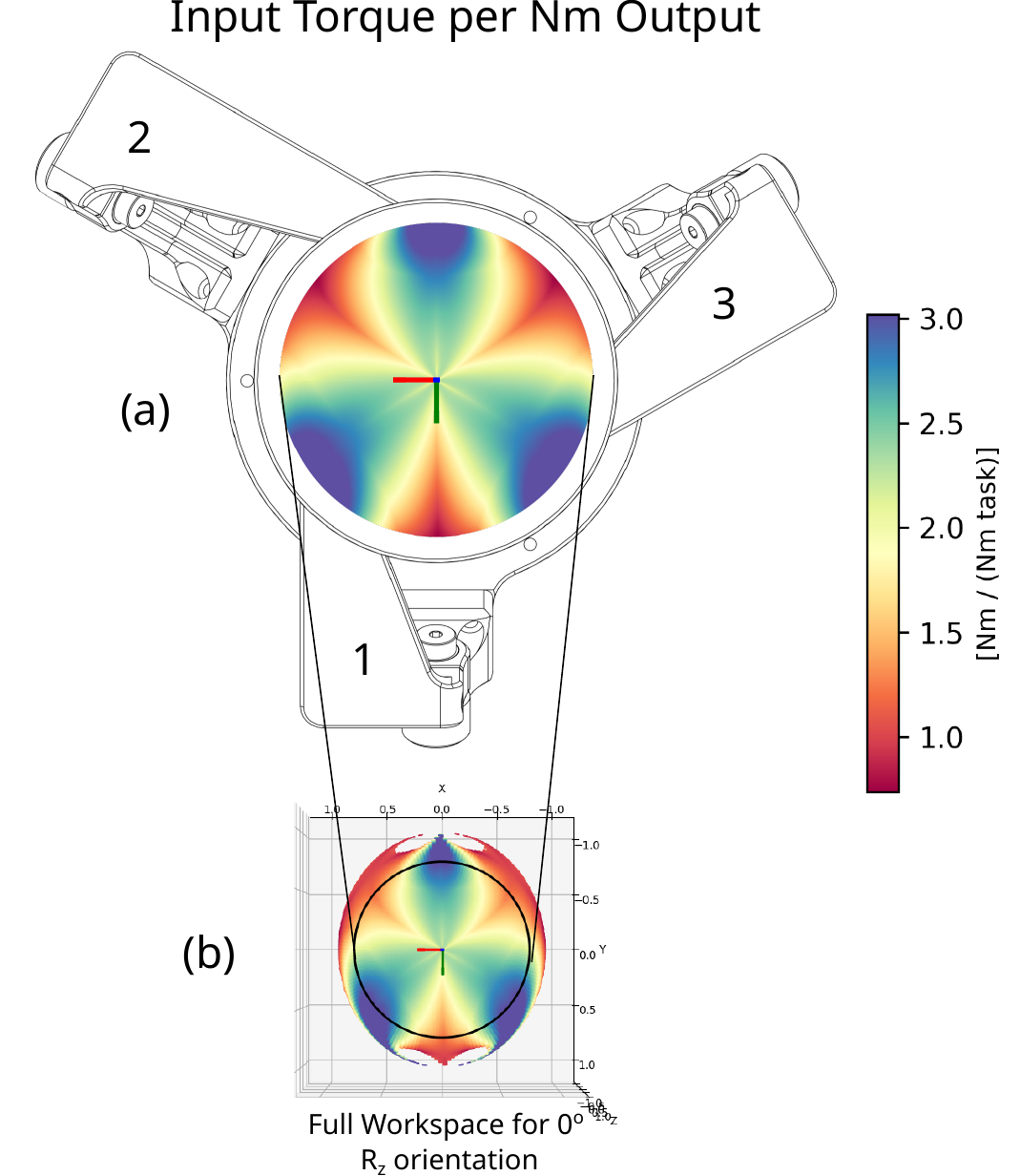}
    \caption{The input to output torque ratio for the SPM across the configuration space is shown for a tendon resistance torque of 1 Nm pulling the joint back to center. The top image shows the maximum output torque given a specific configuration. The configuration limits an individual motor's torque output in the desired direction as described in the Jacobian mapping.}
    \label{fig: spherical_torque_visuals}
\end{figure}

\section{Experiments and Results}

\subsection{Reachable Space}

Taking the Jacobian for the three unique joints, the overall reachable space can be defined using the superposition (addition) of each joint's ROM. A number of these results can be seen in Table \ref{tab:joint_limits}

\subsubsection{Reachable Space of the SPM and Scapular Motion}
The humeral rotation given the initial pose, where the $\mathcal{B}$ and $\mathcal{C}$ frames are aligned, is [+330\degree, -270\degree].
Analysis of the force capabilities and singularities was done and is shown in Figure \ref{fig: spherical_torque_visuals}(a). To avoid singular and dead zones of force, the spherical manipulator's range of motion in the space for flexion and abduction was constrained to a symmetric cone 47\degree rotation in all directions from the center, meaning each has a total range of 94\degree.

\subsubsection{Scapular Motion 4-bar Reachable Space}

The origin of $\mathcal{B}$ in the $\mathcal{A}$ frame moves from (0, .0676, .0564) to (0, .0235, .164) over the range of $q_4$, 23\degree to 70\degree. Giving a total vertical displacement of 10.75 cm and a horizontal displacement of $-4.41$ cm. The scapular angle ranges between 30.72\degree and 145.64\degree, giving an additional 114.92\degree of abduction range to the SPM for a total abduction range of 208.92\degree.

\subsubsection{Pivot Joint}
The final joint is a simple pivot joint and has a range of $\pm 70\degree$. This gives a total of 234\degree of ROM for the shoulder moving forward and backward.

\begin{table}[t!]
\vspace{2.5mm}
\centering
\label{tab:rms_torque_error}
\caption{Error of rendered torque for 9 haptic cones}
\begin{tabular}{|c|c|c|}
\hline
Cone Half-Angle & Stiffness $k$ (Nm/rad) & RMS $|\boldsymbol{\tau}_{xy}|$ Error (Nm) \\
\hline
\multirow{3}{*}{$30^\circ$} & 0.50 & 0.065 \\
                             & 1.00 & 0.086 \\
                             & 1.50 & 0.078 \\
\hline
\multirow{3}{*}{$20^\circ$} & 0.50 & 0.086 \\
                             & 1.00 & 0.066 \\
                             & 1.50 & 0.019 \\
\hline
\multirow{3}{*}{$10^\circ$} & 0.50 & 0.060 \\
                             & 1.00 & 0.028 \\
                             & 1.50 & 0.072 \\
\hline
\end{tabular}
\vspace{-4mm}
\end{table}

\begin{figure*}
\vspace{2.5mm}
    \centering
    \includegraphics[width=\linewidth]{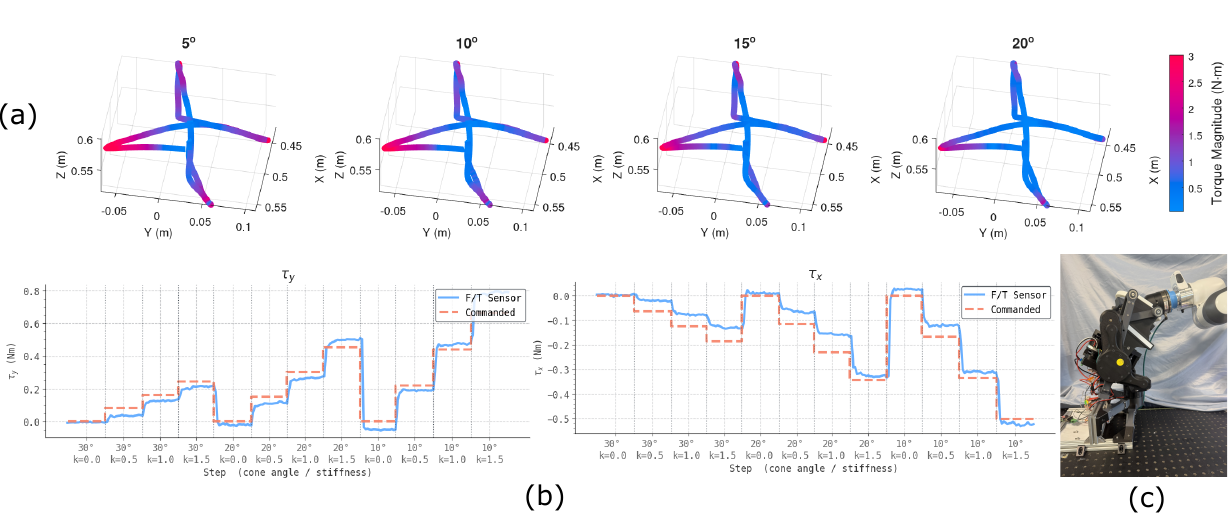}
    \caption{The results of experiments with the Franka Panda arm and ATI Axia80 M8 transducer force-torque sensor. Sub-figure (a) shows the wrench torque perceived by the Panda robot with 4 different shoulder ranges of motion. At the lowest range of motion, the robot perceives the highest resistance forces, emulating the stretching of tendons visualized with a bright pink. The largest range in which the robot does not hit the end ranges during its motion is nearly all shown in blue, indicating very low perceived tendon forces. Sub-figure (b) validates the Jacobian transform between the motor torque and output torque of the device. The ATI Axia80 M8 was used to measure the Haptic wrench (resistance torque), which was compared to the desired input torque direction and magnitude. Sub-figure (c) shows the attachment of the robot to the shoulder with the load cell. }
    \label{fig: experimental_results}
\end{figure*}

\begin{figure*}
    \centering
    \includegraphics[width=1\linewidth]{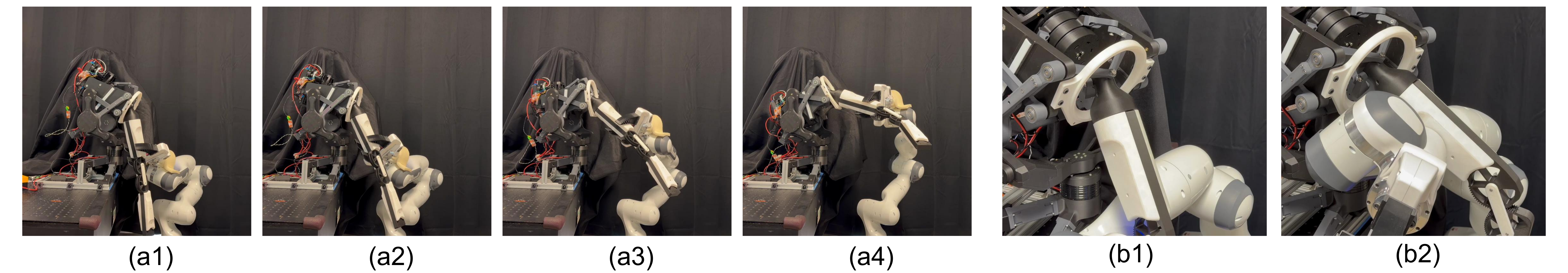}
    \caption{Still captures from experiments involving the use of the Franka Panda platform performing tasks. Series (a) depicts the robot grasping the upper arm and lifting the arm from a relaxed position to a position within an acceptable range for \cite{wang2023one}. The second series (b) shows the arm using WAM to lift the device from the armpit, which can be used to understand motions for patient transfer tasks. }
    \label{fig:fig_of_motion}
    \vspace{-4mm}
\end{figure*}

\subsection{Simulated vs Real Static Force Production}

The force production capabilities of the SPM were calculated in a spherical range of polar angles ranging [0\degree, 47\degree] and azimuthal angles [0\degree, 360\degree]. The end-effector configuration matrix was computed assuming zero yaw (humeral rotation). Inverse kinematics calculated the corresponding joint angles, $q_{1:3}$, that define the Jacobian. Then, the motor torques required to produce 1 Nm of torque to rotate the platform back towards the center are calculated, and the motor torque magnitudes are summed to compare the $\tau_{1:3}$ input to output torque felt at the end effector. Fig \ref{fig: spherical_torque_visuals}(b) shows the production of unconstrained force across the reachable space. Fig \ref{fig: spherical_torque_visuals}(a) shows the simulation of the combined output of the motors working in tandem to provide torque to rotate the arm back to center in the constrained space. As can be seen in the results, there are symmetric regions of overlap where two motors must provide more than 1Nm to produce 1Nm output directed back to the center (dark blue), as well as high torque production regions where motors will not have to input more than 1Nm to get 1Nm out (red). The 4-bar can provide a sustained force of 30 N to the coupler link. The protraction/retraction motor can provide 4 Nm of sustained torque at the joint. 

\subsection{Reach Cone Haptics}

Two experiments were performed to validate the haptic rendering of the SPM. The first validated that the Franka arm could perceive the haptic cone, and the second validated the rendering's accuracy. 
In the first experiment, the Franka arm was directly attached to the shoulder joint and given a star-patterned trajectory, moving the shoulder through flexion/extension and abduction.  Fig \ref{fig: experimental_results} (a) shows the wrench torque perceived by the Franka robot with 4 different shoulder ranges of motion.  With the smallest shoulder range of motion, the robot perceives the highest resistance forces, emulating the stretching of tendons visualized with a bright pink. In the largest range, where the robot just reaches the edge of its motion, its torque perception is nearly all shown in blue, indicating very low perceived tendon forces.

To validate the accuracy of the haptic rendering, static testing was done. The Panda arm was positioned with an ATI F/T sensor, the ATI Axia80 M8 transducer force-torque sensor, between the socket and the Panda end effector, shown in Fig \ref{fig: experimental_results}(c). The haptics were then applied to a set of cones defining the limits of the shoulder joint [$10\degree$, $20\degree$,  $30\degree$] for tendon stiffnesses of [.5 Nm/rad, 1 Nm/rad, 1.5 Nm/rad]. The load cell wrench forces were recorded at this static position across all joint limit cones. The results of these experiments are shown in \ref{fig: experimental_results} (b) and \ref{tab:rms_torque_error}. The error of the rendered torque was between .019 and .086Nm.


\subsection{Robot Experiment}

A second series of experiments were performed using the Franka Emilia Panda Arm to showcase the arm's ability to function as a human phantom for the robot to interact with. One of the resulting trajectories can be seen in Fig \ref{fig:fig_of_motion}. An important trajectory is the trajectory using WAM (whole arm manipulation) reaching under the arm to lift the shoulder, a similar maneuver used in patient transfer tasks and also in the capabilities of the device as seen in Fig \ref{fig:fig_of_motion} (b).

\section{Discussion and Conclusion}

In this paper, we presented a mechanical mechanism with an equivalent motion profile to that of a human shoulder joint. Our experiments demonstrated the device's ability to provide accurate, functional haptic renderings of 3D torques that replicate tendon stiffness at the end ranges of motion, as perceived by a generic manipulation platform. Finally, we presented the function of the device in its intended uses, being manipulated and maneuvered by a generic robotic platform in meaningful contexts: motions critical to in dressing assistance and transfer tasks. 

While the device meets all the basic requirements, there are areas for improvement that would make it easier to work with. Anecdotally, there were issues with the deflection of the rotating members that control the SPM at high force inputs. A future version would use a mechanism to further constrain the SPM's spherical nature under high forces. Additionally, certain design decisions, such as canting the arm segment out of the SPM, could yield even more accurate human joint motion. A final improvement would be to reduce the overall size of the device, as it is both heavy and not a form factor suitable for the shoulder; addressing this could enable cleaner transferability and the use of human pose estimation algorithms in the future.

\balance
\bibliographystyle{ieeetr}
\bibliography{refs}
\balance

\end{document}